\documentclass[11pt,a4paper]{article}
\usepackage[hyperref]{acl2021}
\usepackage{times}
\usepackage{latexsym}

\usepackage{graphicx}
\usepackage{url}
\usepackage{amsmath}
\usepackage{booktabs}
\usepackage{cleveref}
\usepackage{float}
\usepackage{microtype}

\aclfinalcopy % Uncomment this line for the final submission
\def\aclpaperid{3212} %  Enter the acl Paper ID here

\usepackage{wrapfig}
\usepackage{CJK,algorithm,algorithmic,amssymb,amsmath,array,epsfig,graphics,multirow,array,float,subfigure,verbatim,epstopdf}
\usepackage{enumitem}
\usepackage{color,soul}
\usepackage{hhline}
\usepackage{multirow}
\usepackage{hyperref}
\usepackage{tabularx}
\usepackage{bm}
\usepackage{seqsplit}
\usepackage{stmaryrd}
\usepackage{booktabs}
\usepackage[ruled,vlined,algo2e]{algorithm2e}
\usepackage{xparse}

\NewDocumentCommand{\heng}{ mO{} }{\textcolor{red}{\textsuperscript{\textit{Heng}}\textsf{\textbf{\small[#1]}}}}

\NewDocumentCommand{\julia}{ mO{} }{\textcolor{blue}{\textsuperscript{\textit{Julia}}\textsf{\textbf{\small[#1]}}}}

\NewDocumentCommand{\chenkai}{ mO{} }{\textcolor{orange}{\textsuperscript{\textit{Chenkai}}\textsf{\textbf{\small[#1]}}}}

\NewDocumentCommand{\liliang}{ mO{} }{\textcolor{brown}{\textsuperscript{\textit{liliang}}\textsf{\textbf{\small[#1]}}}}

\title{HySPA: Hybrid Span Generation for Scalable Text-to-Graph Extraction}

\author{Liliang Ren, Chenkai Sun,  Heng Ji, Julia Hockenmaier\\ 
  University of Illinois, Urbana Champaign\\
  Department of Computer Science \\
  \texttt{\{liliang3,chenkai5,hengji,juliahmr\}@illinois.edu}
  }

\date{}

\begin{document}
\maketitle
\begin{abstract}
Text-to-Graph extraction aims to automatically extract information graphs consisting of mentions and types from natural language texts. Existing approaches, such as table filling and pairwise scoring, have shown impressive performance on various information extraction tasks, but they are difficult to scale to datasets with longer input texts because of their second-order space/time complexities with respect to the input length. In this work, we propose a \textbf{Hy}brid \textbf{SP}an Gener\textbf{A}tor (\textbf{HySPA}) that invertibly maps the information graph to an alternating sequence of nodes and edge types, and directly generates such sequences via a hybrid span decoder which can decode both the spans and the types recurrently in linear time and space complexities. Extensive experiments on the ACE05 dataset show that our approach also significantly outperforms state-of-the-art on the joint entity and relation extraction task.\footnote{Our code is publicly available at \url{https://github.com/renll/HySPA}
}
\end{abstract}
% \julia{what is "pairwise scoring? perhaps just give a reference? this seems such a generic term} 
\section{Introduction}

Information Extraction (IE) can be viewed as a Text-to-Graph extraction task that aims to extract an information graph \cite{li-etal-2014-constructing,hgraph} consisting of mentions and types from unstructured texts, where the nodes of the graph are mentions or entity types and the edges are relation types that indicate the relations between the nodes. A typical approach towards graph extraction is to break the extraction process into sub-tasks, such as Named Entity Recognition (NER) \cite{florian-etal-2006-factorizing,florian-etal-2010-improving} and Relation Extraction (RE) \cite{sun-etal-2011-semi,jiang-zhai-2007-systematic}, and either perform them separately \cite{chan-roth-2011-exploiting} or jointly \cite{li-ji-2014-incremental,eberts2019span}.% \heng{add more recent work}

Recent joint IE models \cite{dygie,tse,lin-etal-2020-joint}  have shown impressive performance on various IE tasks, since they can mitigate  error propagation  and leverage  inter-dependencies between the tasks. 
Previous work often uses pairwise scoring techniques to identify relation types between entities. However, this approach is computationally inefficient because it needs to enumerate all possible entity pairs in a document, and the relation type is a \emph{null} value for most of the cases due to the sparsity of relations between entities. Also,  pairwise scoring techniques evaluate each relation type independently and thus fail to capture  interrelations between relation types for different pairs of mentions.

\begin{figure}[t]
\centering
\includegraphics[width=.9\columnwidth]{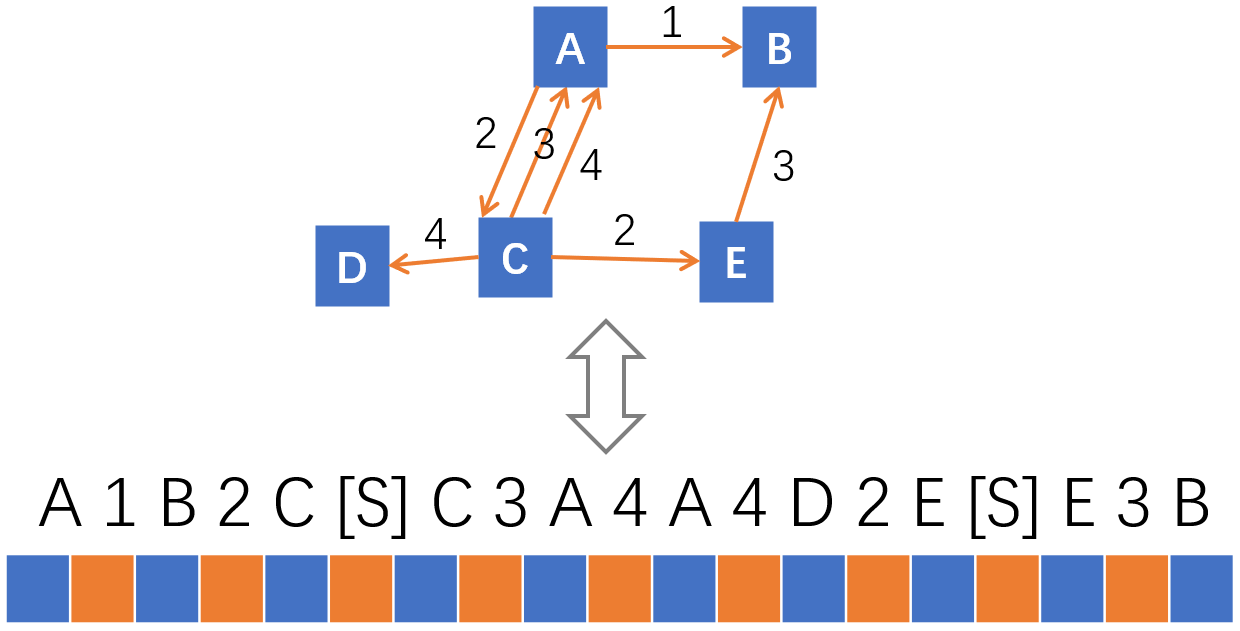}
\caption{We represent directed multigraphs  as \emph{alternating} sequences of nodes (blue) and edges (orange). 
Here, the graph is traversed by Breadth First Search (BFS) with an ascending ordering of nodes and edge types. ``[s]'' or [SEP] is a virtual edge type,  representing the end of each BFS level.}\label{f1}
\end{figure}

Another approach is to treat the joint information extraction task as a table filling problem \cite{zhang-etal-2017-end,tse}, and generate two-dimensional tables with a Multi-Dimensional Recurrent Neural Network \cite{graves2007multi}. This  can capture interrelations among entities and relations, but the space complexity grows quadratically with respect to the length of the input text, making this approach impractical for long sequences.

Some attempts, such as Seq2RDF \cite{liu2018seq2rdf} and IMoJIE \cite{kolluru2020imojie},  leverage the power of Seq2seq models \cite{seq2seq} to capture the interrelations among  mentions and  types with  first-order complexity, but they all use a pre-defined vocabulary for mention prediction, which largely depends on the distribution of the target words and will not be able to handle unseen out-of-vocabulary words. 

To solve these problems, we propose a first-order approach that invertibly maps the target graph to an alternating sequence of nodes and edges, and applies a hybrid span generator that directly learns to generate such alternating sequences. Our main contributions are three-fold:

\begin{itemize}
    \item We propose a general technique to invertibly map between an information graph and an alternating sequence (assuming a given  graph traversal algorithm). Generating an alternating sequence is equivalent to generating the original information graph. 
    \item We propose a novel neural decoder that is enforced to only generate alternating sequences by decoding spans and types in a hybrid manner. For each decoding step, our decoder only has linear space and time complexity with respect to the length of the input sequence, and it can capture inter-dependencies among mentions and types due to its nature as a sequential decision process.
    \item We conduct extensive experiments on the Automatic Content Extraction (ACE) dataset which show that our model achieves state-of-the-art performance on the joint entity and relation extraction task which aims to extract a knowledge graph from a piece of unstructured text. %\heng{looks like you did not do event extraction. drop it from here. but you did relation extraction on both sentence-level and document-level, you can try to highlight that here.}
\end{itemize}

\section{Modeling Information Graphs as Alternating Sequences}

An \textbf{information graph} can be viewed as a heterogeneous multigraph \cite{li-etal-2014-constructing,hgraph} $G=(V,E)$, where $V$ is a set of nodes (typically representing spans $(t_s, t_e)$ in the input document) and $E$ is a multiset of edges with a node type mapping function $\phi: V \rightarrow Q$ and an edge type mapping function $\psi: E \rightarrow R$. Node and edge types are  assumed to be drawn from a finite vocabulary. Node types can be used e.g. to represent entity types (PER, ORG, etc.), while edge types may represent relations (PHYS, ORG-AFF, etc.) between the nodes.
%The function $\phi$ is an injection, \emph{i.e.}, each node $v_i\in V$ has only one node type. 
In this work, we represent node types as separate nodes that are connected to their node $v$ by a special edge type, [TYPE]. \footnote{$Q$ includes a [NULL] node type for the case when the input text does not have an information graph.}
%\julia{You never talk about NULL anywhere else -- footnote?} 

\paragraph{Representing information graphs as sequences}
Instead of directly modeling the  space of heterogeneous multigraphs, $\mathcal{G}$, we  build a mapping  $s^{\pi} =f_s(G,\pi)$ from  $\mathcal{G}$, to a sequence space $S^\pi$. $f_s$  depends on a (given) ordering $\pi$ of nodes and their edges in $G$, constructed by a graph traversal algorithm like Breadth First Search (BFS) or Depth First Search (DFS), and an internal ordering of nodes and edge types.
We assume that the elements $s^{\pi}_i$ of the resultant sequences $s^{\pi}$ are drawn from finite sets of node representations $V$ (defined below), node types $Q$, edge types $R$ (incl. [TYPE]), and  ``virtual" edge types $U$: $ \forall~s_i^{\pi}\in  ~s^{\pi}, ~s_i ^{\pi}\in V \cup Q \cup R \cup U$.  Virtual edge types $U =\{ \text{[SOS]}, \text{[EOS]}, \text{[SEP]}\}$  do not represent edges in $G$, but serve to control the generation of the sequence, indicating the start/end of sequences and the separation of levels in the graph.

We furthermore assume that  $s^\pi=s^\pi_0,...,s^\pi_n$ that represent graphs have an \textbf{alternating} structure, where $s^\pi_0, s^\pi_2,s^\pi_4,...$ represent nodes $V$, and $s^\pi_1,s^\pi_3,...$  represent actual or virtual edges.  In the case of BFS, we exploit the fact that it visits nodes level by level, \emph{i.e.}, in the order $\text{p}_{i}, \text{c}_{i1},...,\text{c}_{ik}, \text{p}_{j}$ (where $c_{ik}$ is the $k$-th child of parent $p_i$, connected by edge $e_{ik}$, and $\text{p}_j$  may or may not be equal to one of the children of $\text{p}_i$), which we turn into a sequence,
\begin{align*}
    s^\pi = \text{p}_{i},& \psi(\text{e}_{i1}),  \text{c}_{i1},...,\\
    &\psi(\text{e}_{ik}),\text{c}_{ik} ,\text{[SEP]},\text{p}_{j},...
\end{align*}
where we use the special edge type [SEP] to delineate the levels in the graph. This representation allows us to unambiguously recover the original graph, if we know which type of graph traversal is assumed (BFS or DFS).\footnote{In the case of DFS, [SEP] tokens appear after leaf nodes. Parents appear once for each child.} 
Algorithm 1 (which we use to translate graphs in the training data to sequences) shows how an alternating sequence for a given graph can be constructed with BFS traversal.  \Cref{f1} shows the alternating sequence for an information multigraph. 
 The length $|s^{\pi}|$ is bounded linearly by the size of the graph $O(|s^{\pi}|)=O(|V|+|E|)$ (which is also the complexity of typical graph traversal algorithms like BFS/DFS).

 \begin{algorithm}[H]
\SetAlgoLined
\SetKwInOut{Input}{Input}\SetKwInOut{Output}{Output}
\Input{Ordered adjacency dictionary of an information graph $G$, positions of nodes in the input text $p_q$, frequency of edge types in the training set $p_r$}
%\julia{I don't understand the roles of $p_q$ vs. $p_r$ here -- do you want to show that you can use both, or either? this is unclear}
%\liliang{information graph is a multigraph, so the edges also need to be sorted}

 \Output{An alternating sequence $y^\pi$}
 
\BlankLine

Sort the nodes in $G$ according to $p_q$

For each node $v$ in $G$, sort the neighbors and the edges of $v$ according to $p_q$ and $p_r$ respectively

Instantiate $y^\pi$ as an empty list

 \For{$u$ in $G$}{
\If{$u$ is not visited}{
    Initialize an empty queue $q$\;
    
    Mark $u$ as visited and enqueue $u$ to $q$ \;
    
   \While{$q$ is not empty}{
   Dequeue a node $w$ from $q$\;
   
   \If{$w$ in $G$}{
    Append $w$ and all the neighbors of $w$ with their edge types to $y^\pi$\;
    
    Append the separation edge type, [SEP], to $y^\pi$\;
    
    Mark all unvisited neighbors of $w$ as visited and enqueue them to $q$\;
   }
   }
}
}
 \textbf{Return} $y^\pi$
 \caption{Alternating sequence construction algorithm with BFS}
\end{algorithm}
% \julia{Liliang -- do you want to remove the following,  and perhaps just move the GraphRNN reference somewhere else? I'm still not sure about the bijection, because it's only a bijection under $\pi$, and it doesn't matter in practice. What matters in practice is that we can recover graphs from sequences.}
% For all sequences $s^\pi \in S^\pi$, the mapping $f_s$ is automatically a bijection, since the ordering $\pi$ enforces the uniqueness of $s^\pi$ as the representation of graph $G$, similar to the case of a simple graph \cite{graphrnn}. Once the ordering $\pi$ is given, one can unambiguously recover $G$ from $s^\pi$.\julia{You don't need the ordering for recovery}.

%To solve this problem, we define the \emph{alternating} sequence, $y^\pi$, as a special type of sequence whose elements alternate between nodes and edge types. \Cref{f1} shows an example of the alternating sequence for an information multigraph. Given the ordering $\pi$, we build $y^\pi$ by appending the separation edge type, [SEP], at the end of each traversal level when constructing the sequences $s^\pi \in S^\pi$ from the graph traversal algorithm, $A$. The magnitude of 

\paragraph{Node and Edge Representations}
Our node and edge representations (explained below) rely on the observation that there are only two kinds of objects in an information graph: spans (as addresses to pieces of input texts) and types (as representations of abstract concepts). Since we can view types as special spans of length 1 grounded on the vocabulary of all types, $Q\cup R \cup U$, we only need $O(nm+|Q\cup R \cup U|)$ number of indices to unambiguously represent the spans grounded on a concatenated representation of the type vocabulary and the input text, where $n$ is the maximum input length, $m$ is the maximum span length, and  $m\ll n$. We denote these indices as \emph{hybrid spans} because they consist of both the spans of texts and the length-1 spans of types. These indices can be invertibly mapped back to types or text spans depending on their magnitudes (details of this mapping are explained in Section \ref{map}). With this joint indexing of spans and types, the task of generating an information graph is thus converted to generating an alternating sequence of \emph{hybrid spans}.

% epresent spans $t=(t_s,t_e) \in \mathbb{N}^2, t_s<t_e$ whose maximum length $m$ is significantly smaller than the maximum input length $n$. That is, we only have most $m$ different span lengths, and up to $n$ different starting positions, and hence we only need an inventory of $O(nm)$ indices to unambiguously represent spans. 

% \julia{Liliang: I think the rest of this section is just what I had copied from the decoder part. Do you wnat to leave this here?}

\paragraph{Generating sequences}
We model the distribution $p(s^{\pi})$ by a sequence generator $h$ with parameters $\theta$ ($|s^{\pi}|$ is the length of the $s^{\pi}$):
\begin{align*}
        p(s_{i}^{\pi} | s_{0}^{\pi}, ... ,s_{i-1}^{\pi}) &= h(s_{0}^{\pi}, ... ,s_{i-1}^{\pi},\theta),\\
        p(s^{\pi}) &= \prod_{i=1}^{|s^{\pi}|} p(s_{i}^{\pi} | s_{0}^{\pi}, ... ,s_{i-1}^{\pi}),
\end{align*}
We will address in the following sections how to enforce the sequence generator, $h$, to only generate  sequences in the space  $S^\pi$, since we do not want $h$ to assign non-zero probabilities to  arbitrary sequences that do not have a corresponding graph. 

\begin{figure*}[htb]
\centering
\includegraphics[width=13cm]{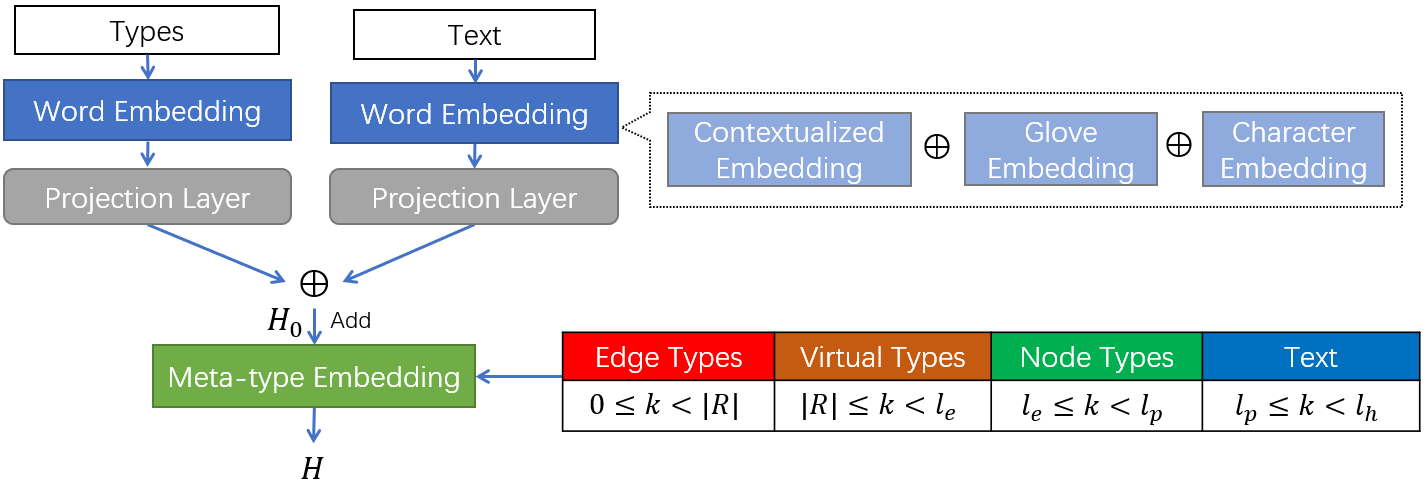}
\caption{The encoder architecture of our model, where the $\oplus$ symbol is the concatenation operator, $k$ is the index of the word vectors in $H_0$, and $l_e=|R|+|U|$. The colored table on the right indicates the assignment of the meta-types for different blocks of the concatenated word vectors from $H_0$. }\label{f2}
\end{figure*}
% Naively applying the mapping $f_s$ to graph $G$ will be sub-optimal since $f_s$ is a one-to-many mapping and the sequence generator, $h$, may not learn the equivalencies between different sequences observed from the same graph and harm its generalization power. 

% Contrary to the previous works that solely enforce the orderings of the sequence obtained from the mapping $f_s$, we propose to also enforce the orderings during the generation step of the sequence generator, $h$

\section{HySPA: Hybrid Span Generation for Alternating Sequences}

In order to directly generate a target  sequence that alternates between  nodes that represent spans in the input and a set of node/edge types that depend on our extraction task, we first build a hybrid representation $H$ that is a concatenation of the hidden representations from edge types, node types and the input text. This representation functions as both the context space and the output space for our decoder. Then we invertibly map both the spans of input text and the indices of the types to the \emph{hybrid spans} grounded on the representation $H$.
Finally, hybrid spans are generated auto-regressively through a hybrid span decoder to form the alternating sequence $y^\pi \in S^\pi$. By translating the graph extraction task to a sequence generation task, we can easily use beam-search decoding to reduce possible exposure bias \cite{beam} of the sequential decision process and thus find globally better graph representation.

\paragraph{High-level overview of HySPA:} The HySPA model takes a piece of text (e.g. a sentence or passage), and the pre-defined node and edge types as input, and outputs an alternating sequence representation of an information graph. We enforce the generation of this sequence to be alternated by applying an alternating mask to the output probabilities. The detailed architecture is described in the following subsections.

\subsection{Text and Types Encoder}

\Cref{f2} shows the encoder architecture of our proposed model. For the set of node types, $Q$, and the set of edge types, $R$, and the virtual edge types, $U$, we arrange the type list, $\mathbf{v}$ as a concatenation of the label names of the edge types, virtual edge types and node types, \emph{i.e.}, 
% a given piece of text, $\mathbf{x}$,
\begin{align*}
     \mathbf{v}&=\hat{R} \oplus \hat{U} \oplus \hat{Q}\\
\hat{R} &= [R_1,...,R_{|R|}]\\
\hat{U} &= [U_1,...,U_{|U|}]\\
\hat{Q} &= [Q_1,...,Q_{|Q|}]
\end{align*}
where $\oplus$ means the concatenation operator between two lists, and $\hat{R},\hat{U},\hat{Q}$ are the lists of the type names in the sets $R,U,Q$, respectively (e.g. $\hat{Q} = [\text{``Geopolitics''}, \text{``Person''},...]$). Note that the concatenation order between the lists of type names can be arbitrary as long as it is kept consistent throughout the whole model.
Then, as in the embedding part of the table-sequence encoder~\cite{tse}, for each type, $v_i$, we embed the label tokens of the types with the contextualized word embedding from a pre-trained language model, the GloVe embedding \cite{glove} and the character embedding,
%\julia{what are the words of the type? Do you mean the type labels? if so, how do you use a contextualized representation?}
%we first append a special task token ``$\langle$Task$\rangle$'', at the beginning of the sequence to indicate the specific task types for the encoder, \emph{i.e.}, the ``$\langle$Task$\rangle$'' token can be ``$\langle$EvE$\rangle$'', ``$\langle$RE$\rangle$'' and ``$\langle$EnE$\rangle$'' to represent event extraction, relation extraction and entity extraction tasks respectively. We then take the word embedding and position embedding of the input sequence, $x'$, and sum them together. We may also optionally add the co-reference or type embedding as prior knowledge for 
%julia{are you concatenating vectors into a matrix, or low-dimensional vectors into one high-dimensional vector? perhaps use a different symbol for the two different concatenation operations?} 

%\julia{there is no  ordering in a set... you mean that since the types form a set we assign to each of them an arbitrary index in the type embedding matrix?}
%Note that the ordering of the types in a set can be arbitrary since we do not add position embeddings between the types, and the following self-attention layers are naturally permutation equivariant. 
\begin{align*}
E_1 &=\text{ContextualizedEmbed}(\mathbf{v}), \in R^{~l_p \times d_c }\\
E_2 &=\text{GloveEmbed}(\mathbf{v}), \in R^{~l_p \times d_g }\\
E_3 &=\text{CharacterEmbed}(\mathbf{v}), \in R^{~l_p \times d_k } \\
E_4 &= E_1 \oplus E_2 \oplus E_3 \in R^{~l_p \times d_e},\\
E_v & = E_4 W_0^T \in R^{~l_p \times d_m},
\end{align*} 
where $l_p=|R|+|U|+|Q|$ is the number of all kinds of types, $W_0\in R^{~d_e \times d_m}$ is the weight matrix of the linear projection layer, $d_e = d_c+d_g+d_k$ is the total embedding dimension and $d_m$ is the hidden size of our model. After we obtain the contextualized embedding of the tokens of each type $v_i\in \mathbf{v}$, we take the average of these token vectors as the representation of $v_i$ and freeze its update during training. More details of the embedding pipeline can be found in Appendix \ref{hyper}.

%\julia{I don't fully understand the following sentence}
This embedding pipeline is also used to embed the words in the input text, $\mathbf{x}$. Unlike the pipeline for the type embedding, we represent the word as the contextualized embedding of its first sub-token from the pre-trained Language Model (LM, e.g. BERT \cite{bert}), and finetune the LM in an end-to-end fashion.

% \julia{I don't understand the role that $E_v$ plays in the encoder. or rather, if I do, I don't think it's described well here. or perhaps I don't actually  understand $H_0$}
After obtaining the type embedding $E_v$, and the text embedding $E_x$ respectively, we concatenate them along the sequence length dimension to form the hybrid representation $H_0$. Since $H_0$ is a concatenation of word vectors from four different types of tokens, \emph{i.e.}, edge types, virtual edge types, node types and text, a meta-type embedding is applied
 to indicate this type difference between the blocks of vectors from the representation $H_0$, as shown in \Cref{f2}. The final context representation $H$ is obtained by element-wise addition of the meta-type embedding and $H_0$,
\begin{align*}
    H_0 &= E_v \oplus E_x \in R^{l_h \times d_m},\\
    H_s & = \text{MetaTypeEmbed}(H_0) \in R^{l_h \times d_m},\\
    H&= H_0 + H_s\in R^{l_h \times d_m},
\end{align*}
where $l_h =l_p+|x|$ is the height of our hybrid representation matrix $H$.

 \subsection{Invertible Mapping between Spans \& Types and Hybrid Spans}\label{map}

\begin{figure}[t]
\centering
\includegraphics[width=\columnwidth]{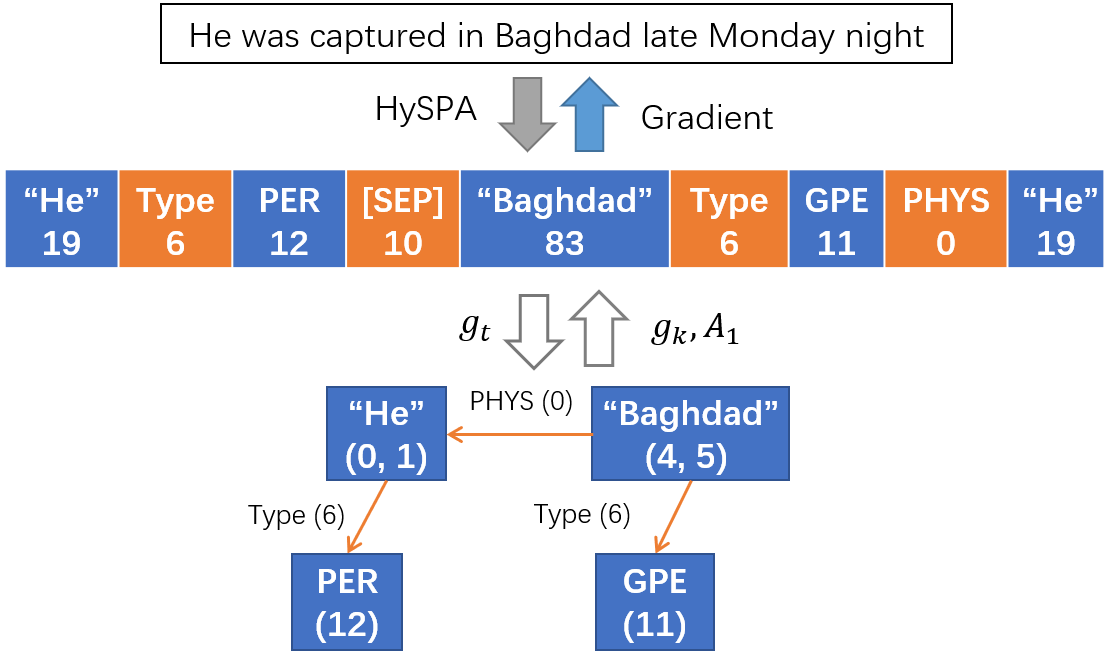}
\caption{An example of the alternating sequence representation (in the middle) of a knowledge graph (at bottom) from the ACE05 training set, where $A_1$ means the Algorithm 1. We take $m =16$ and $l_p =19$ for this example. ``19" in the alternating sequence is the index for the span (0,1) of ``He", ``83" is the index for the span (4,5) of ``Baghdad", and  ``10" is the index of the virtual edge type, [SEP]. The input text (on top) for this graph is ``He was captured in Baghdad late Monday night".}\label{f4}
\end{figure}

Given a span in the text, $t=(t_s,t_e) \in \mathbb{N}^2, t_s<t_e$, we convert the span $t$ to an index $k$, $k\geq l_p$, in the representation $H$ via the mapping $g_k$,
\begin{equation*}
k=g_k(t_s,t_e) = t_s m+t_e-t_s-1+l_p \in \mathbb{N},
\end{equation*}
where $m$ is the maximum length of spans, and $l_p=|R|+|U|+|Q|$. We keep the type indices in the graph unchanged because they are smaller than $l_p$ and $k\geq l_p$. Since, for an information graph, the maximum span length, $m$, of a mention is often far smaller than the length of the text, \emph{i.e.}, $m\ll n$, we can then reduce the bound of the maximum magnitude of $k$ from $O(n^2)$ to $O(nm)$ by only considering spans of length smaller than $m$, and thus maintain  linear space complexity for our decoder with respect to the length of the input text, $n$. \Cref{f4} shows a concrete example of our alternating sequence for a knowledge graph in the ACE05 dataset.

Since $t_s,t_e,k$ are all natural numbers, we can construct an inverse mapping $g_t$ that converts the index $k$ in $H$ back to $t=(t_s,t_e)$,
\begin{align*}
t_s=g_{t_s}(k)=-\text{max}&(0,-k+l_p)+\\
    &\lfloor \text{max}(0,k-l_p)/m \rfloor +l_p ,\\
t_e=g_{t_e}(k)=g_{t_s}(k)&+\text{max}(0, k-l_p) ~\text{mod}~m,
\end{align*}
where $\lfloor \cdot \rfloor$ is the integer floor function and $\mod$ is the modulus operator.  Note that $g_t(k)$ can be directly applied to the indices from the types segment of $H$ and remain their values unchanged, \emph{i.e.},  
$$
g_t(k)=(k,k), \forall k<l_p, k\in \mathbb{N}.
$$
With this property, we can easily incorporate the mapping $g_t$ into our decoder to map the alternating sequence $y^\pi$ back to the spans in the hybrid representation $H$.
% Since a span is a tuple of start position $t_s$ and end position $t_e$ with the constraint, $t_s< t_e$, that cannot be easily beam-searched, we need to create a mapping to invertibly convert a span to an index in the context representation $H$ to support beam-search over the space of the alternating sequence.
% \julia{Putting this back in, since this hasn't been addressed. is the problem just beam search, or the fact that a tuple is two tokens that break the alternating sequence assumption (and have the constraint that one is smaller than the other?)? I'm also not sure I understand what you mean by "cannot be directly beam-searched".}
% \julia{Putting this back in, since this hasn't been addressed. I think you need to explain how you represent spans  much earlier in the paper. I was wondering earlier how you'd represent a sequence of tokens with your alternating representation}

\subsection{Hybrid Span Decoder}

%%\julia{the width was 16cm, but I changed it to 15 to save space}
\begin{figure*}[htb]
\centering
\includegraphics[width=14cm]{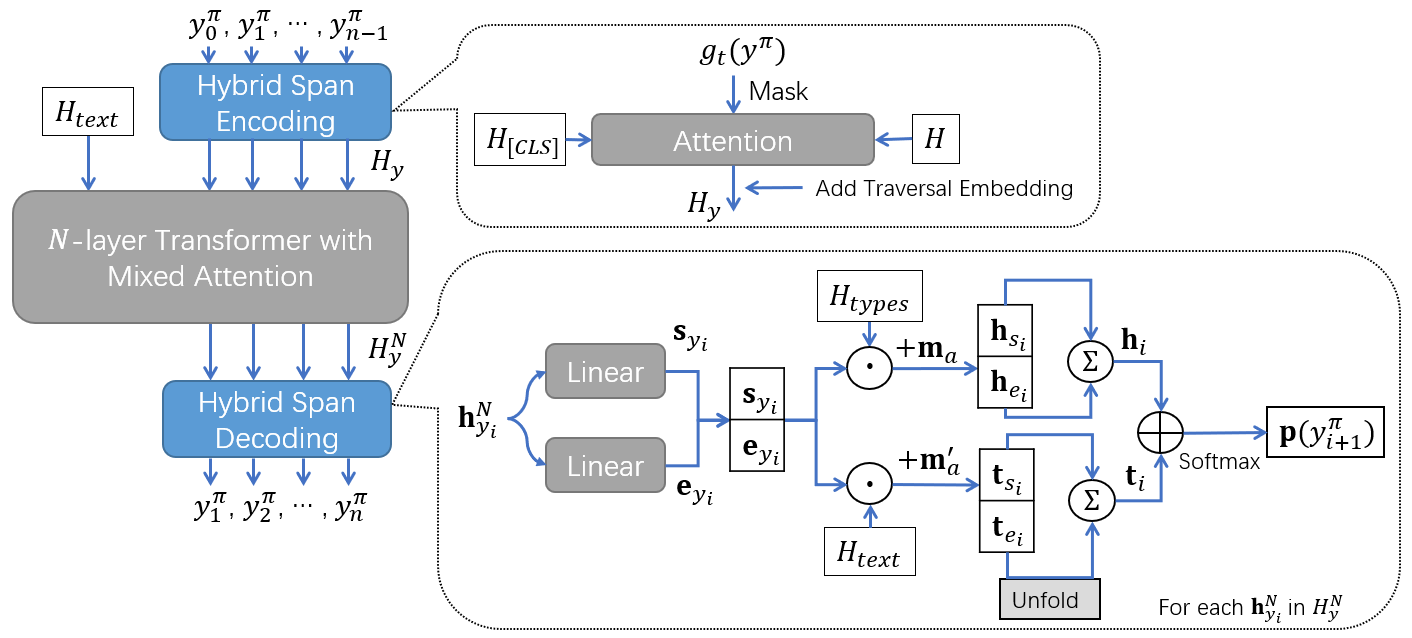}
\caption{The  architecture of our hybrid span decoder. $N$ is the number of the decoder layers. $\oplus$ before the softmax function means the concatenation operator. $H_y^N$ is the hidden representation of the sequence $y^\pi$ from the last decoder layer. Our hybrid span decoder can be understood as an auto-regressive model that operates in a closed context space and output space defined by $H$.}\label{f3}
\end{figure*} %

\Cref{f3} shows the general model architecture of our hybrid span decoder. Our decoder takes the context representation $H$ as input, and recurrently decodes the alternating sequence $y^\pi$ given a start-of-sequence token.

%  \julia{Liliang, part of the problem is that the variables in the text don't all appear in Figure 4}
 
% \paragraph{$H_{\text{text}}$} 
% \julia{what is $H_{\text{text}}$, and how does it relate to the hybrid span encoding part?} 

\begin{figure}[t]
\centering
\includegraphics[width=\columnwidth]{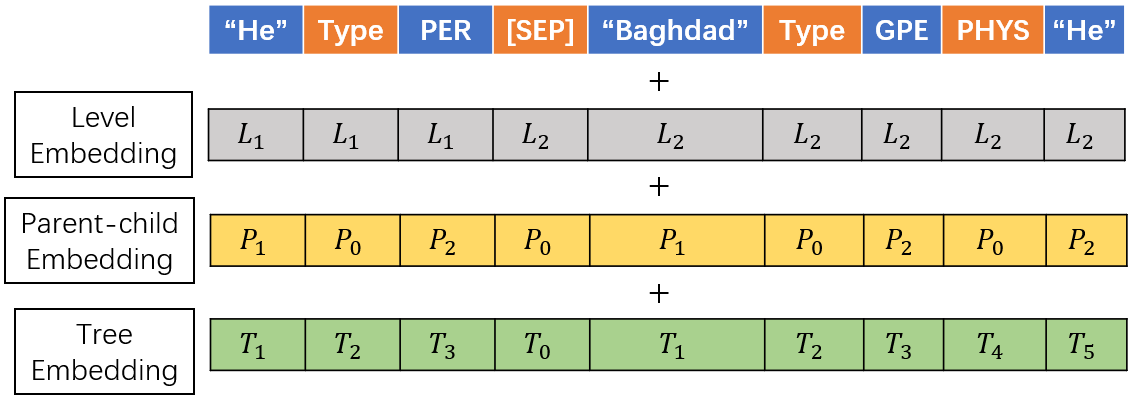}
\caption{An example of BFS traversal embedding for an alternating sequence, [``He'', Type, PER, [SEP], ``Baghdad'', Type, GPE, PHYS, ``He'']. Our traversal embedding is the sum of the level embedding, the parent-child embedding and the tree embedding.}\label{ft}
\end{figure}

\paragraph{Hybrid Span Encoding via Attention}
% \julia{Can you first give an intuition of what this component does at a high level?}
Given the alternating sequence $y^\pi$, and the mapping $g_t$ (section \ref{map}), our decoder first maps each index in $y^\pi$ to a span, $(t_{s_i}, t_{e_i})=g_t(y_i^\pi)$, grounded on the representation $H$ and then converts the span to an attention mask, $M_0$, to allow the model to learn to represent a span as a weighted sum of a segment of the contextualized word representations referred by the span,
\begin{align*}
    Q & =W_1^TH_{[CLS]}+\mathbf{b}_1 \quad \in R^{~|y^\pi|\times d_m},\\
    K & =W_2^TH+\mathbf{b}_2 \quad \in R^{~l_h \times d_m},\\
    H_y &=\text{softmax}\bigg(\frac{QK^T}{\sqrt{d_m}}+M_0
     \bigg)H \quad \in R^{~|y^\pi|\times d_m},\\
  M_0&(i,j)=\left\{
\begin{array}{c l}	
    0,  & t_{s_i}\leq j \leq t_{e_i}\\
     -\infty, & \text{otherwise}
\end{array}\right.     
\end{align*}
where $H_{[CLS]}\in R^{~|y^\pi|\times d_m}$ is the $|y^\pi|$-times repeated hidden representation of the start of the sequence token, [CLS], from the text segment of $H$, and $H_y$ is our final representation of the hybrid spans in $y^\pi$. $W_1, W_2,\mathbf{b}_1,\mathbf{b}_2$ are learnable parameters, and $t_{s_i}, t_{e_i}$ are the start and the end position of the span thatwe are encoding. Note that for the type spans whose length is 1, the result of the \emph{softmax} calculation will always be 1, which leads to its span representation to be exactly its embedding vector as we desired.

\paragraph{Traversal Embedding}
In order to distinguish the hybrid spans at different position in $y^\pi$, a naive way is to
add a sinusoidal position embedding \cite{tran} to $H_y$. However, this approach treats the alternating sequence as an ordinary sequence and ignores the underlying graph structure it encodes. To alleviate this issue, we propose a novel traversal embedding approach which captures the traversal level information, the parent-child information and the intra-level connection information as a substitution of the naive position embedding. Our traversal embedding can either encode the BFS or DFS traversal pattern. As an example, we assume BFS traversal here and leave the details of DFS traversal embedding in Appendix \ref{dfst}.

Our BFS traversal embedding is a pointwise sum of the level embedding, $L$, the parent-child embedding, $P$, and the tree embedding, $T$ of a given alternating sequence, $y$,
$$
\text{TravEmbed}(y) = L(y)+P(y)+T(y) \in R^{~|y|\times d_m}
$$
 where the level embedding assigns the same embedding vector $L_i$ for each position at the BFS traversal level $i$, and the value of the embedding vector is filled according to the non-parametric sinusoidal position embedding since we want our embedding to extrapolate to the sequence that is longer than any sequences in the training set. The parent-child embedding assigns different random initialized embedding vectors at the positions of the parent nodes and the child nodes in the BFS traversal levels to help model distinguish between these two kinds of nodes. For encoding the intra-level connection information, our insight is that the connection between each nodes in a BFS level can be viewed as a depth-3 tree, where the first depth takes the parent node, the second depth is filled with the edge types and the third depth consists of the corresponding child nodes for each of the edge types. Our tree embedding is then formed by encoding the position information of the depth-3 tree with a tree positional embedding \cite{treepos} for each BFS level. \Cref{ft} shows a concrete example of how these embeddings function for a given alternating sequence. The obtained traversal embedding is then pointwisely added to the hidden representation of the alternating sequence $H_y$ for injecting the traversal information of the graph structure.

% Following \citet{tran}, add a sinusoidal position embedding to $H_y$ to distinguish between each time stamp of the sequential decision process.

\paragraph{Inner blocks}

With the input text representation $H_{text}$ sliced from the hybrid representation $H$ and the target sequence representation $H_y$, we apply an $N$-layer transformer structure with mixed-attention \cite{layerwise} to allow our model to utilize features from different attention layers when decoding the edges or the nodes of an alternating sequence. Note that our hybrid span decoder is perpendicular to the actual choice of the neural structures of the inner blocks, and we choose the design of mixed-attention transformer \cite{layerwise} because its layerwise coordination property is empirically more suitable for our heterogeneous decoding of two different kinds of sequence elements. The detailed structure of the inner blocks is explained in Appendix \ref{mix}.

\paragraph{Hybrid span decoding}

For the hybrid span decoding module, we first slice off the hidden representation of the alternating sequence $y^\pi$ from the output of the $N$-layer inner blocks and denote it as $H_y^N$.
Then for each hidden representation $\mathbf{h}_{y_i}^N\in H_y^N,  0\leq i < |y^\pi|$, we apply two different linear layers to obtain the start position representation, $\mathbf{s}_{y_i}$, and the end position representation, $\mathbf{e}_{y_i}$,
\begin{align*}
    \mathbf{s}_{y_i} & =W_5^T\mathbf{h}_{y_i}+\mathbf{b}_5 \quad \in R^{~ d_m},\\
    \mathbf{e}_{y_i} & =W_6^T\mathbf{h}_{y_i}+\mathbf{b}_6 \quad \in R^{~ d_m},
\end{align*}
where $W_5,W_6\in R^{~d_m \times d_m}$ and $\mathbf{b}_5,\mathbf{b}_6 \in R^{~d_m}$ are learnable parameters.
Then we  calculate the scores of the target spans separately for the types segment and the text segment of $H$, and concatenate them together before the final softmax operator for a joint estimation of the probabilities of text spans and type spans,
\begin{align*}
    \mathbf{h}_{s_i} & =  H_{\text{types}} ~\mathbf{s}_{y_i}  +\mathbf{m}_a & &\in R^{l_p},\\
      \mathbf{h}_{e_i}& =  H_{\text{types}} ~\mathbf{e}_{y_i} +\mathbf{m}_a & &\in R^{l_p},\\
      \mathbf{h}_{i}& = \mathbf{h}_{s_i}+\mathbf{h}_{e_i}& &\in R^{l_p},\\
    \mathbf{t}_{s_i} & =  H_{\text{text}} ~\mathbf{s}_{y_i}  +\mathbf{m}_a'& &\in R^{n},\\
      \mathbf{t}_{e_i}& =  H_{\text{text}} ~\mathbf{e}_{y_i} +\mathbf{m}_a'& &\in R^{n}, \\ 
    \mathbf{t}_i &=  \text{unfold}(\mathbf{t}_{e_i},m)+\mathbf{t}_{s_i}& &\in R^{nm},\\
    \mathbf{p}(y_{i+1}^\pi) &= \text{softmax}( \mathbf{h}_{i} \oplus \mathbf{t}_i )& &\in R^{nm+l_p},
\end{align*} 
where 
$\mathbf{h}_i$ is the score vector of possible spans in the type segment of $H$, and $\mathbf{t}_i$ is the score vector of possible spans in the text segment of $H$. Since the type spans always have a span length 1, we only need an element-wise addition between the start position scores, $\mathbf{h}_{s_i}$ and the end position scores $\mathbf{h}_{e_i}$ to calculate $\mathbf{h}_i$. The entries of $\mathbf{t}_i$ contain the scores for the text spans, $t_{s_{i},j}+t_{e_{i},k}, \forall j\leq k, k-j < m$, which are calculated with the help of an \emph{unfold} function which converts the vector $\mathbf{t}_{e_i}\in R^n$ to a stack of $n$ sliding windows of size $m$, the maximum span length, with stride 1. The alternating masks $\mathbf{m}_a \in R^{l_p},\mathbf{m}_a' \in R^n$ are defined as:
\begin{align*}
  \mathbf{m}_a(j)=&\left\{
\begin{array}{c l}	
    0, &  y_i^\pi > l_e \cap j < l_e \\
     -\infty, & \text{otherwise}
\end{array}\right. \\  
  \mathbf{m}_a'(j)=&\left\{
\begin{array}{c l}	
    -\infty, &  y_i^\pi > l_e \\
     0, & \text{otherwise}
\end{array}\right.   
\end{align*}
 where $l_e = |R|+|U|$ is the total number of  edge types.  In this way, while we have a joint model of  nodes and edge types, the output distribution is enforced by the alternating masks to produce an alternating decoding of  nodes and  edge types, and this is the main reason why we call this decoder a hybrid span decoder.

%\julia{I still get completely lost in this section. Also, please explain unfold better. What dimensionality does $\mathbf{t}_{e_i}$ have? }

\begin{table*}
\centering
\begin{tabular}{lllll}
\hline
\textbf{IE Models} & \textbf{Space Complexity} & \textbf{Time Complexity}  &\textbf{NER} & \textbf{RE} \\
\hline
PointerNet \cite{ptr} & $O(n)$ &$O(n^2)$ & 82.6 & 55.9 \\
SpanRE \cite{dixit-al-onaizan-2019-span} & $O(n)$ &$O(n^2)$   & 86.0 & 62.8\\
Dygie++ \cite{dygie} & $O(n)$ &$O(n^2)$ & 88.6 & 63.4 \\
OneIE \cite{lin-etal-2020-joint} & $O(n)$ &$O(n^2)$ &88.8 & 67.5 \\
TabSeq \cite{tse} & $O(n^2)$ &$O(n)$ & 89.5 & 67.6 \\
\hline
HySPA (ours) \quad w/ RoBERTa  & \multirow{2}{4em}{$O(n)$} &\multirow{2}{4em}{$O(n)$}& 88.9 & \textbf{68.2} \\
\quad \quad \quad \quad\quad \quad\quad w/ ALBERT &  & & \textbf{89.9} & \textbf{68.0} \\
\hline
\end{tabular}
\caption{\label{res}
Joint NER and RE F1 scores of the IE models on the ACE05 test set.  Complexities are calculated for the entity and relation decoding part of the models ($n$ is the length of the input text). The performance of the TabSeq model reported here is based on the same ALBERT-xxlarge \cite{albert} pretrained language model as ours.
}
\end{table*}

\section{Experiments}

\subsection{Experimental Setting}

We test our model on the ACE 2005 dataset distributed by LDC\footnote{\url{https://catalog.ldc.upenn.edu/LDC2006T06}},  %\heng{put LDC catalog instead of citation for this data set} \chenkai{Do you mean put a footnote like this?}
which includes 14.5k sentences, 38.3k entities (with 7 types), and 7.1k relations (with 6 types), derived from the general news domain. More details can be found in Appendix \ref{data}.

Following previous work, we use F1 as an evaluation metric for both NER and RE. For the NER task, a prediction is marked correct when both the type and the boundary span match those of the gold entity. For the RE task, a prediction is correct when both the relation type and the boundaries of the two entities are correct.

% For document-level relation extraction, since the output is a sequence of spans (and relation ids), \heng{what do you mean by ids?} which often doesn't match an entity mention's span exactly, we use Jaccard similarity coefficient  (intersection cardinality / union cardinality) to measure matching degree between span and entity mention  \heng{"span and entity mention" is confusing. do you mean "gold-standard mentions and system extracted mentions"?} based on the set of indices of each. For example, entity 1 spans from 1 to 6 (inclusively), entity 2 spans from 1 to 7, entity 3 spans form 0 to 4, and the predicted span is 0 to 6, then the matching scores between the predicted span and the three entities will respectively be, 6/7,  6/8, and 5/7. In this case, the predicted span will be assigned as entity 1 because they have the highest matching score. \heng{this paragraph is very confusing. maybe draw a figure to illustrate it}  \heng{this metric is complicated and weird. Why don't you use the standard way to evaluate relation extraction? just check head 1, head 2 and type. you can check the metric in Ying's paper}

\subsection{Implementation Details}
When training our model, we apply the cross-entropy loss with a label smoothing factor of 0.1. The model is trained with 2048 tokens per batch (roughly a batch size of 28) for 25000 steps using an AdamW optimizer \cite{adamw} with a learning rate of $2e^{-4}$, a weight decay of 0.01, and an inverse square root scheduler with 2000 warm-up steps. Following the TabSeq model \cite{tse}, we use RoBERTa-large \cite{roberta} or ALBERT-xxlarge-v1 \cite{albert} for the pretrained language model and slow its learning rate by a factor of 0.1 during training. A hidden state dropout rate of 0.2 is applied to RoBERTa-large while the rate of 0.1 for  ALBERT-xxlarge-v1. A dropout rate of 0.1 is also applied to our hybrid span decoder during training. We set the maximum span length, $m=16$, the hidden size of our model, $d_m=256$, and the number of the decoder blocks, $N=12$. Even though theoretically the beam-search should help us reduce the exposure bias, we do not observe any performance gain during grid search of the beam size and the length penalty on the validation set (detailed grid search setting is in Appendix \ref{hyper}). Thus we set a vanilla beam size of 1 and the length penalty of 1, and leave this theory-experiment contradiction for future research. Our model is built with the FAIRSEQ toolkit \cite{ott2019fairseq} for efficient distributed training and all the experiments are conducted on two NVIDIA TITAN X GPUs.

\subsection{Results}

\Cref{res} compares our model with the previous state-of-the-art results on the ACE05 test set. Compared with the previous SOTA, TabSeq \cite{tse} with ALBERT pretrained language model, our model with ALBERT has significantly better performance for both NER score and RE score, while maintaining a linear space complexity which is an order smaller than TabSeq. Our model is the first joint model that has both linear space and time complexities compared with all previous joint IE models, and thus has the best scalability for large-scale real world applications.

\subsection{Ablation Study}

\begin{table}[t!]
\begin{center}
\begin{tabular}{lcc} 
\toprule \bf Model & \bf NER F1 & \bf RE F1 \\ \midrule
HySPA  w/ RoBERTa &88.9 &68.2  \\
~ -- Traversal-embedding  &88.9 &66.7 \\
\quad -- Masking  &88.1 &64.8 \\
\quad -- BFS  &88.7 &66.2  \\
\quad -- Mixed-attention  &88.6 &64.7  \\
\quad -- Span-attention  &88.5 &66.1 \\
\bottomrule
\end{tabular}
\end{center}
\caption{\label{abm} Ablation study on the ACE05 test set. ``-- Traversal-embedding'': we remove the traversal embedding and instead use sinusoidal position embedding, and the following ablations are based on the model after this ablation. ``-- Masking'':  we remove the alternating mask from the  hybrid span decoder. ``-- BFS": we use DFS instead of BFS as traversal. ``-- Mixed-attention":  we remove the mixed-attention layer and use a standard transformer encoder decoder structure. ``-- Span-attention":  we remove the span attention in the span encoding module and instead average the words in the span.}
\end{table}

To prove the effectiveness of our approach, we conduct ablation experiments on the ACE05 dataset. As shown in \Cref{abm}, after we remove the traversal embedding the RE F1 scores drop significantly, which indicates that our traversal embedding can help encode the graph structure and improve relation predictions. Also if the alternating masking is dropped, the NER F1 and RE F1 scores both drop significantly, which proves the importance of enforcing the alternating pattern. 
We can observe that the mixed-attention layer contributes significantly for relation extraction. This is because the layer-wise coordination can help the decoder to disentangle the source features and utilize different layer features between the entity and the relation prediction. We can also observe that the DFS traversal has worse performance than BFS. We suspect that this is because the resultant alternating sequence from DFS is often longer than the one from BFS due to the nature of the knowledge graphs, and thus increases the learning difficulty.

% For ``- ShareParam'', we remove the parameter sharing mechanism on the encoders and the attention module. For ``- Order'', we further arrange the order of the slots according to its global frequencies in the training set instead of the local frequencies given the domain it belongs to. For ``- Nested'', we do not generate domain sequences but generate \emph{combined slot} sequences which combines the domain and the slot together. For ``- BlockGrad'', we further remove the gradient blocking mechanism in the CMR decoder

\subsection{Error Analysis}
\label{section:analysis}

\begin{figure}[t]
\centering
\includegraphics[width=\columnwidth]{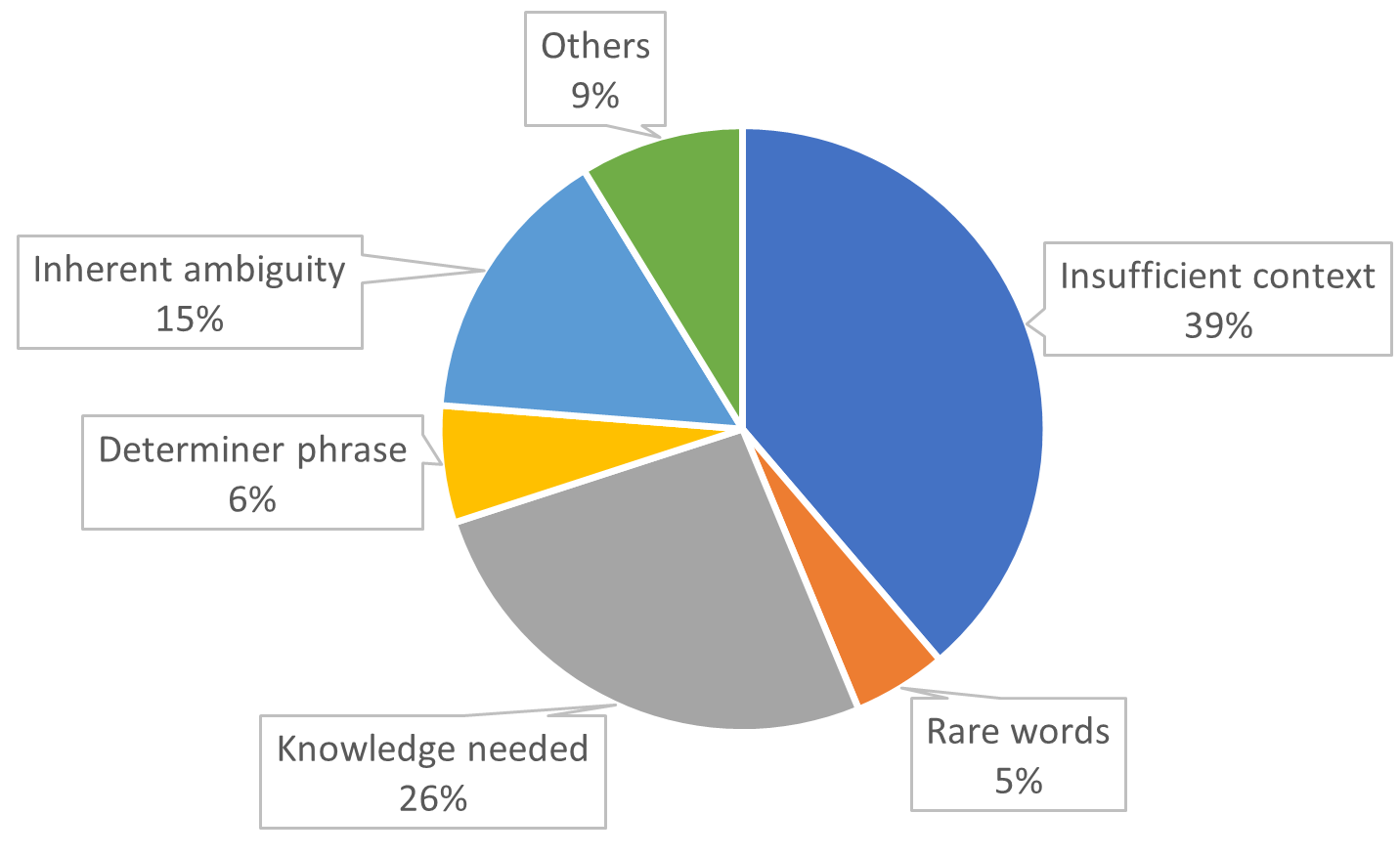}
\caption{Distribution of remaining errors on the ACE05 test set.}
\label{edist}
\end{figure}

After analyzing 80 remaining errors, we categorize and discuss common cases 
below (Figure~\ref{edist} plots the distribution of  error types). These  may require additional features and strategies to address. 

%\noindent Shared deficit

\noindent \textbf{Insufficient context}. In many examples, the answer entity is  a  pronoun that cannot be accurately typed given the limited  context: in \textit{``We notice they said they did not want to 
use the word destroyed, in fact, they said let others do that''}, it's 
difficult to correctly classify  \textit{We} as an 
organization. 
This  could be mitigated by using entire documents as input, leveraging cross-sentence context. 

% We also noticed that, often when entity typing is incorrect, the relation could be consequently misclassified as well or not detected. 
% Another solution would be to learn jointly models coreference between entities among different examples.

\noindent \textbf{Rare words}. The rare word issue is when the word in test 
set rarely 
appeared in the training set and often not termed in the dictionary. In the sentence \textit{``There are 
also Marine FA-18s and 
Marine Heriers at this base''}, the term \textit{Heriers} (a vehicle incorrectly classified as person by the model) 
neither
appeared in the training set, nor understood well by pre-trained language 
model; the model, in this case, can only rely on subword-level representation.

% \noindent  \textbf{Coreference}. Another issue arises when our model has a hard time 
% identifying non-pronoun coreference 
% %\heng{what do you mean by 'non-pronoun coreference'} 
% as the entity. Common examples include \textit{some of}, \textit{most of}, and \textit{many of}. The problem could be 
% potentially alleviated by incorporating syntactic information (e.g., dependency 
% parsing) to identify the grammatical structure of the sentence.
%\julia{or by doing coreference first?}
%\julia{Do you mean the model has difficulties identifying the entity type of pronominal mentions? "some of" is not really a pronoun, and wouldn't stand on its own. Do you mean it can't parse "some of X" as one noun phrase?}
%\heng{maybe you can say that in the future you can apply entity linking to acquire some external knowledge}

% \textcolor{red}{Determiner phrase detection. Another issue arises when our model has a hard time identifying/classifying determiner phrases (DP) entities in the partitive construction. For example in the sentence ``..., many of the people refuse to eat their Charms'',
% our model doesn't predict the entity of ``many''. The problem could be potentially alleviated by incorporating syntactic information (e.g., dependency 
% parsing) to identify the grammatical structure of the sentence}

% Common examples include \textit{some of}, \textit{most of}, and \textit{many of}. The problem could be 

\noindent \textbf{Background knowledge required} Often the sentence mentions entities that are difficult to infer from the context, but are easily identified by consulting a knowledge base: in \textit{``but critics say Airbus should have sounded a stronger alarm after a similar incident occurred in 1997''}, our model incorrectly predicts the \textit{Airbus} to be a vehicle while the \textit{Airbus} here refers to the European aerospace corporation. Our system also separated \textit{United Nations Security Council} into two entities \textit{United Nations} and \textit{Security Council}, generating a non-existing relation triple (\textit{Security Council} part-of \textit{United Nations}). Such mistakes could be avoided by consulting a knowledge base such as DBpedia~\cite{dbpedia} or by performing entity linking. 
% involving ``Sheikh Hamad bin Jassem bin Jabr 
% al-Thani'', our 
% system classifies the whole phrase as an entity, while Sheikh is fact a 
% language specific address 
% instead of being part a name. By linking to the page of ``Hamad bin Jassem 
% bin Jabr al-Thani'', the 
% entity could be better handled. 

%
%Also in the sentence ``There should be an investigation but they should look 
%into the 
%circumstances , ' said Lance Corporal Christopher Hanson . ' "", our system 
%detects lance corporal as a name the Lance 
%Corporal is a rank instead of name, 
%----sample 603 ----
%
%sentence: central command says the iraqis shot down an a-10 tank killer today .
%
%pred_entities_sample [(['central', 'command'], 'ORG'), (['iraqis'], 'PER'), 
%(['a-10'], 'VEH'), (['tank'], 'VEH'), (['killer'], 'PER')]
%sample['entities']   [(['central', 'command'], 'ORG'), (['iraqis'], 'PER'), 
%(['killer'], 'VEH')
%
%----sample 1223 ----
%
%sentence: There should be an investigation but they should look into the 
%circumstances , " said Lance Corporal Christopher Hanson . "
%
%pred_entities_sample [(['they'], 'ORG'), (['Corporal'], 'PER'), 
%(['Christopher', 'Hanson'], 'PER')]
%sample['entities']   [(['they'], 'PER'), (['Lance', 'Corporal'], 'PER'), 
%(['Christopher', 'Hanson'], 'PER')]

\noindent \textbf{Inherent ambiguity} Many examples have inherent ambiguity, e.g.   \textit{European Union} can be typed as organization or 
political entity, while some entities (e.g., military bases) can be both locations and organizations, or facilities.
%\julia{what do you mean by "have a single type for all entities? do you mean all mentions of that entity? why would this confuse the learning system?}

% \subsubsection{Document-Level Relation Extraction}
% A common error made by the model is as following\newline

% \noindent \textit{Gold: ...12007 18 514...}

% \noindent \textit{Prediction: ...2219 18 5142 18 5142 18 5142 18 5142 18 5142 (18 5142)*22 ..}\newline

% \noindent where there are excessively repeating sub-sequence.

% \subsubsection{Sentence-Level Relation Extraction}
% The first type of error is that the model is unable to identify entity with numerical suffix well. For example, the model made the following mistakes:\newline

% \noindent \textit{Prediction 1: (['587'], 'VEH')}

% \noindent \textit{Gold 1: (['flight 587'], 'VEH’)}

% \noindent \textit{Prediction 2: (['flight'], 'VEH'),
% (['903'], 'VEH’)}

% \noindent \textit{Gold 2: (['flight 903'], 'VEH’)}\newline

% The second type of error is that the model cannot do co-reference very well, for instance for the sentence "...the only way to get that kind of legitimate government to have it look good...", where "it" a co-reference of government, the model did not predict "it". This means adding co-reference training data or use a co-reference oriented pre-trained language model might improve the result.

% \subsection{Qualitative Analysis}
% \heng{add error analysis here, show error category and examples}

\section{Related Work}

NER is often done jointly with RE in order to mitigate error propagation and learn inter-relation between tasks. One line of approaches is to treat the joint task as a squared table filling problem~\cite{tse_prior, tse_prior2, tse}, where the $i$-th column or row represents the $i$-th token. The table has diagonals indicating sequential tags for entities and other entries as relations between pairs of tokens. Another line of work is by performing RE after NER. In the work by~\citet{bilstm_joint}, the authors used BiLSTM~\cite{bilstm_original} for NER and consequently a Tree-LSTM~\cite{treelstm} based on dependency graph for RE. \citet{dygie} and \citet{dygie_original}, on the other hand, takes the approach of constructing dynamic text span graphs to detect entities and relations. Extending on \citet{dygie}, \citet{lin-etal-2020-joint} introduced \textsc{One}IE, which further incorporates global features based on cross subtask and instance constraints, aiming to extract IE results as a  graph. Note that our model differs from \textsc{One}IE~\cite{lin-etal-2020-joint} in that our model captures global relationships automatically through autoregressive generation while \textsc{One}IE uses feature engineered templates; 
Moreover, \textsc{One}IE needs to do pairwise classification for relation extraction, while our method efficiently generates existing relations and entities. 

While several Seq2Seq-based models~\cite{seq2umtree, seq2seq_re, seq2seq_re2, seq2seq_re3,zhang-etal-2019-broad} have been proposed to generate triples (i.e., node-edge-node), our model is fundamentally different from them in 
that: (1) it is
generating a BFS/DFS traversal of the target graph, which captures 
dependencies between nodes and edges and has a shorter target sequence, (2) we model the nodes as the spans in the text, which is independent of the vocabulary, so even if the tokens of the nodes are rare or unseen words, we can still generate spans on them based on the context information.

% Our work, unlike any other methods, performed both task at the same time by generating entities and relations from the text, eliminating the need for sequence labeling and pairwise classification.

% , which has not been. Given the promising performance and efficiency, we believe, this could open up new direction for joint , or even general information extraction task

% EE

\section{Conclusion}
In this work, we propose the Hybrid Span Generation (HySPA) model, the first end-to-end text-to-graph extraction model that has a linear space and time complexity at the graph decoding stage. Besides its scalability, the model also achieves  state-of-the-art performance on the ACE05 joint entity and relation extraction task. Given the flexibility of the structure of our hybrid span generator, abundant future research directions remain, e.g. incorporating the external knowledge for hybrid span generation, applying more efficient sparse self-attention, and developing better search methods to find more globally plausible graphs represented by the alternating sequence.

\section*{Acknowledgments}
This work is supported by Agriculture and Food Research Initiative (AFRI) grant no. 2020-67021-32799/project accession no.1024178 from the USDA National Institute of Food and Agriculture.
\bibliographystyle{acl_natbib}
\bibliography{anthology,acl2021}

\appendix

\section{Hyperparameters}\label{hyper}
We use 100-dimensional GloVe word embeddings trained on 6B tokens as intialization \footnote{\url{https://nlp.stanford.edu/projects/glove/}}, and freeze its update during training. The character embedding has 30-dimension with LSTM encoding \footnote{\url{https://github.com/LorrinWWW/two-are-better-than-one/blob/master/layers/encodings/embeddings.py}} and the Glove Embeddings for the out of vocabulary tokens are replaced with randomly initialized vectors following \citet{tse}. We use gradient clipping of 0.25 during training. The number of heads for our mixed attention is set to 8. The beam size and length penalty is decided by a grid-search on the validation set of the ACE05 dataset, and the range for the beam size is from 1 to 7 with a step size of 1 and the length penalty is from 0.7 to 1.2 with a step size of 0.1. We choose the best beam size and length penalty based on the metric of relation extraction F1 score.

\section{Training Details}\label{training}

Our model has 236 million parameters with the ALBERT-xxlarge pretrained language model. On average, our best performing model with ALBERT-xxlarge can be trained distributedly on two NVIDIA TITAN X GPUs for 20 hours.

\section{Data} \label{data}
The Automatic Content Extraction (ACE) 2005~\footnote{\url{https://www.ldc.upenn.edu/collaborations/past-projects/ace}} dataset contains English, Arabic and Chinese training data for the 2005 Automatic Content Extraction (ACE) technology evaluation, providing entity, relation, and event annotations. We follow~\citet{dygie}~\footnote{\url{https://github.com/dwadden/dygiepp/tree/master/scripts/data/ace05/preprocess}} for preprocessing and data splits. The preprocessed data contains 7.1k relations, 38k entities, and 14.5k sentences. The split contains 10051 samples for training, 2424 samples for development, and 2050 for testing.

\section{DFS Traversal Embedding}\label{dfst}

Since the parent-child information is already contained in the intra-level connections of DFS traversal, we only have the sum of the level embedding and the connection embedding for DFS traversal embedding. Similar to BFS embedding, the DFS level embedding assigns the same embedding vector $L_i$ for each position at the DFS traversal level $i$, but the value of the embedding vector is randomly initialized instead of filled with the non-parametric sinusoidal position embedding, since the proximity information does not exist between the traversal levels of DFS. However, we do have clear distance information for the elements in a DFS level, \emph{i,e.}, for a DFS level $D =[\text{A, B, C, ..., [sep]} ] $, the distance from A to the elements [A, B, C, ..., [sep]] is $[0, 1, 2, 3, ..., |D|-1]$. We encode this distance information with the sinusoidal position embedding which becomes our connection embedding that captures the intra-level connection information.

\section{Transformer with Mixed-attention}\label{mix}

\begin{figure}[t]
\centering
\includegraphics[width=\columnwidth]{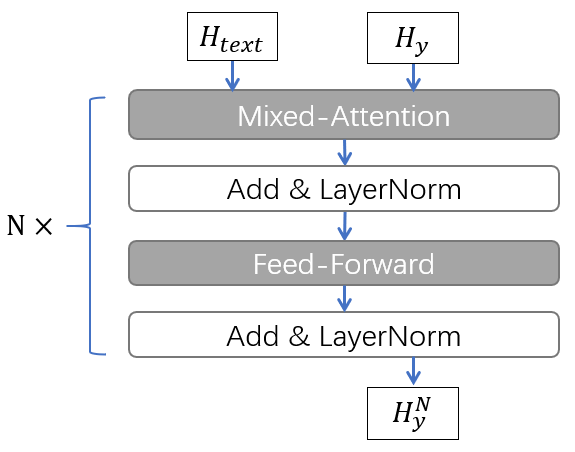}
\caption{The general model architecture of the mixed-attention transformer.}\label{fm}
\end{figure}

 We first slice off the hidden representation of the input text from the hybrid representation $H$, and denote it as $H_\text{text}$, then the input text representation $H_{\text{text}}$ and the output from the Hybrid Span Encoding $H_y$ gets fed into a stack of $N$ mixed-attention/feedforward blocks that have the following structure (as shown in \Cref{fm}):

Since generating the node and  edge types may need  features from different layers, we use mixed attention \cite{layerwise}, which allows our model to utilize the features from different attention layers when encoding the text  segment, $H_{text}$, and the target features, $H_y$,
\begin{align*}
    \text{MixedAtt}(Q,K,V)&=\text{softmax}\bigg(\frac{QK^T}{\sqrt{d_m}}+M_1\bigg)V \\
  &\quad \quad\in R^{~l_m\times d_m},\\
  M_1(i,j)=&\left\{
\begin{array}{c l}	
    0, &j<n \cup j\leq i+n\\
     -\infty, & \text{otherwise}
\end{array}\right.
\end{align*}
where $n=|x|$ is the length of the input text, $l_m=|x|+|y^\pi|$ is the total length of the source and the target features. Denoting the concatenation of the source features, $H_\text{text}$, and the target features, $H_y$, as $H_0$, a source/target embedding \cite{layerwise} is also added to $H_0$ before the first layer of the mixed attention to allow the model to distinguish the features from the source and the target sequences. The mixed-attention layer is  combined with a feed-forward layer to form a decoder block:
\begin{align*}
    \text{FFN}&(x) = \text{max}(0,xW_3+b_3)W_4+b_4,\\
    & Q =W_q^TH_0+\mathbf{b}_q,\\
    & K =W_k^TH_0+\mathbf{b}_k,\\
     &V  =W_v^TH_0+\mathbf{b}_v,\\
    H_0' &=\text{LayerNorm}(\text{MixedAtt}(Q,K,V)+H_0),\\
    H_1 &=\text{LayerNorm}(\text{FFN}(H_0')+H_0'),
\end{align*}
where $W_{q,k,v}, \mathbf{b}_{q,k,v}, W_3\in R^{~d_m\times 4 d_m},W_4\in R^{~4d_m\times d_m},\mathbf{b}_3,\mathbf{b}_4 $ are the learnable parameters, and LayerNorm is the Layer Normalization layer \cite{LayerN}. The decoder block is stacked $N$ times to obtain the final hidden representation $H_N$, and output the final representation of the target sequence, $H_y^N$. The mixed-attention has a time complexity of $O(n^2)$ when encoding the source features, but we can cache the hidden representation of this part when generating the target tokens due to the causal masking of the target features, and thus maintain a time complexity of $O(n)$ for each decoding step.

\end{document}